\documentclass[dvipsnames,format=sigconf,anonymous=false,review=false]{acmart}
\usepackage{algorithm}
\usepackage{algpseudocode}
\usepackage{color}
\usepackage{graphicx}
\usepackage{subcaption}  
\usepackage{amsmath}

\usepackage{amssymb,mathrsfs}
\usepackage{booktabs}
\usepackage{newtxtext,newtxmath}
\usepackage{float}
\AtBeginDocument{%
  \providecommand\BibTeX{{%
    \normalfont B\kern-0.5em{\scshape i\kern-0.25em b}\kern-0.8em\TeX}}}

\setcopyright{acmcopyright}
\copyrightyear{2024}
\acmYear{2024}
\acmDOI{10.1145/1122445.1122456}

\acmConference[GECCO '24]{Genetic and Evolutionary Computation Conference (GECCO '24)}{July 14--18, 2024}{Melbourne, Australia}
\acmBooktitle{Genetic and Evolutionary Computation Conference (GECCO '24),
  July 14--18, 2024, Melbourne, Australia}
\acmPrice{00.00}
\acmISBN{978-1-4503-XXXX-X/18/06}

\begin{document}

\title{SAIS: A Novel Bio-Inspired Artificial Immune System Based on Symbiotic Paradigm}


\author{Junhao Song}
\authornote{Both authors contributed equally to this research.}
\affiliation{%
  \department{Faculty of Engineering}
  \institution{Imperial College London}
  \city{London}
  \country{United Kingdom}
}
\email{junhao.song23@imperial.ac.uk}

\author{Yingfang Yuan}
\authornotemark[1]
\affiliation{%
  \department{School of Mathematical and Computer Sciences}
  \institution{Heriot-Watt University}
  \city{Edinburgh}
  \country{United Kingdom}
}
\email{y.yuan@hw.ac.uk}

\author{Wei Pang}
\authornote{Corresponding author.}
\affiliation{%
  \department{School of Mathematical and Computer Sciences}
  \institution{Heriot-Watt University}
  \city{Edinburgh}
  \country{United Kingdom}
}
\email{w.pang@hw.ac.uk}
\renewcommand{\shortauthors}{Junhao and Yingfang, et al.}

\begin{abstract}
 We propose a novel type of Artificial Immune System (AIS): Symbiotic Artificial Immune Systems (SAIS), drawing inspiration from symbiotic relationships in biology. SAIS parallels the three key stages (i.e., mutualism, commensalism and parasitism) of population updating from the Symbiotic Organisms Search (SOS) algorithm. This parallel approach effectively addresses the challenges of large population size and enhances population diversity in AIS, which traditional AIS and SOS struggle to resolve efficiently. We conducted a series of experiments, which demonstrated that our SAIS achieved comparable performance to the state-of-the-art approach SOS and outperformed other popular AIS approaches and evolutionary algorithms across 26 benchmark problems. Furthermore, we investigated the problem of parameter selection and found that SAIS performs better in handling larger population sizes while requiring fewer generations. Finally, we believe SAIS, as a novel bio-inspired and immune-inspired algorithm, paves the way for innovation in bio-inspired computing with the symbiotic paradigm.
\end{abstract}




\maketitle

\section{Introduction}
With the swift advancement of artificial intelligence, bio-inspired algorithms have become crucial in tackling complex optimisation challenges \cite{vikhar2016ea, brownlee2011clever, eiben2015ec}. In this field, Artificial Immune Systems (AIS) drawing inspiration from biological immune mechanisms, offers a distinctive approach to addressing optimisation problems. Many AIS algorithms conceptualize antigens as the objectives or constraints of a problem, while antibodies are viewed as potential solutions. These AIS systems dynamically update the antibodies to better align with the antigens (identifying optimal solutions), mirroring the adaptive mechanisms of the biological immune system \cite{brownlee2011clever}. Despite the wide-ranging potential applications of AIS across various fields, contemporary research on AIS has predominantly concentrated on the refinement of individual antibodies \cite{brownlee2007clonal, ji2007nsa, kur2000imnet, hal2007symbais}. As highlighted in \cite{hal2007symbais}, there is a notable lack of emphasis on the interactions among antibodies within existing AIS algorithms. This limited scope of research restricts the capabilities of AIS in tackling complex optimisation challenges. This is particularly evident in scenarios involving the generation of antibodies at a large scale \cite{brownlee2011clever, tim2004ais}, where traditional AIS algorithms struggled to deal with challenges like loss of population diversity and low computational efficiency \cite{zheng2010aso, segel2001design}.

This research presents the Symbiotic Artificial Immune System (SAIS), an innovative AIS algorithm employed to address the aforementioned limitations. Inspired by the symbiotic relationships in biology, SAIS not only focuses on interconnections among antibodies, but also enhances their collective problem-solving capabilities. We parallelized three stages (mutualism, commensalism and parasitism) of the Symbiotic Organisms Search (SOS) algorithm \cite{cheng2014sos} to SAIS. This parallel application significantly boosts the algorithm's efficiency and effectiveness in addressing complex optimisation challenges. Through comparative experiments involving multiple algorithms, including SAIS and SOS in 26 distinct benchmark problems, we showcase the results achieved by SAIS in optimising solutions. These findings substantiate the effectiveness of symbiotic interactions among antibodies, thereby affirming the validity of our approach.

The rest of this paper is structured as follows. Section \ref{sect2} introduces the symbiotic relationship in the biological world and existing research of the AIS. Section \ref{sect3} details the algorithmic design and implementation of SAIS. In Section \ref{sect4}, we evaluate the performance of SAIS through a series of benchmark tests under varying parameters, and we conduct comparative analysis with other algorithms. Furthermore, we provide an in-depth discussion of the experimental results that elucidate the advantages of SAIS. Section \ref{sect5} evaluates the role of different population relationships in SAIS by removing different population relationships (mutualism, commensalism and parasitism) from the symbiotic relationship in SAIS. Finally, Section \ref{sect6} summarizes the entire study and proposes future research directions for SAIS.

\section{Related Work} \label{sect2}

\subsection{Symbiotic Relationships in Biology}
The word 'Symbiotic' is derived from the Greek, meaning 'to live together'. It was first used by De Bary in 1878 to describe the act of living together between different organisms (species) \cite{sapp1994evolution, cheng2014sos}. Biology has discovered in the last five years that symbiotic relationships are also influencing the human immune system. For example, probiotics in the human body can help humans digest and increase immunity \cite{ck2019foundation, belkaid2014mic}. Symbiotic relationship may be positive, negative, or neutral. There are three basic types of symbiosis: mutualism, commensalism, and parasitism \cite{ck2019foundation}. Mutualism benefit both species. Commensalism can be one species benefits and the other species is not affected. However, parasitism can benefit one species and harm another. 

\begin{figure}[h]
  \centering
  \includegraphics[width=0.7\linewidth]{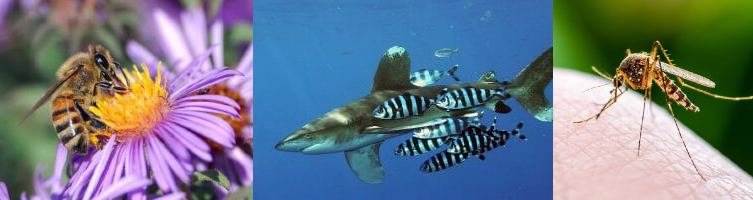}
  \caption{Examples of Mutualism, Commensalism and Parasitism in Biology (CC BY-ND 2.0 Licenses)}
  \Description{Mutualism between flower and bee, commensalism between shark and Remora fish, malaria-transmitting mosquitoes infect human}
  \label{fig_symbexamples}
\end{figure}

As shown in Figure \ref{fig_symbexamples}, bees collect nectar as food and help flowers spread pollen. This mutualism relationship not only promotes their respective survival and reproduction, but also has a positive impact on the health and diversity of the entire ecosystem \cite{ck2019foundation}. The relationship between shark and Remora fish is a typical example of commensalism. The Remora fish attaches to the shark's body and uses the shark's movements to swim and find food more easily, without the shark itself being significantly affected. In the rightmost of Figure \ref{fig_symbexamples}, mosquitoes transmitting malaria to humans is an example of a parasitic relationship. This mosquito ingests blood by biting humans, thereby transmitting malaria and posing a serious threat to human health. Mosquitoes gain benefits, but humans suffer negative consequences \cite{ck2019foundation, rooks2016gut}.

The study of symbiotic relationships in biology reveals complex interaction mechanisms. These mechanisms play a key role in maintaining biodiversity and ecological balance. These findings provide inspiration for simulating interactions in nature in the field of computational intelligence.

\subsection{Research on the AIS and Symbiotic AIS Applications} \label{subsect_symbAIS_SOS}
In the field of computational intelligence, Artificial Immune Systems (AIS) is a collective name for a variety of algorithms inspired by biological immune systems \cite{tim2004ais, brownlee2011clever}. Many AIS simulate the dynamic interactions between antibodies (potential solutions to a problem) and antigens (the target solutions of the problem), though specific implementations like the B cell algorithm \cite{Kelsey2003bcell} and dendritic cell algorithm \cite{greensmith2008dendritic} may not directly use antibodies. While algorithms for AIS have been extensively studied, most existing methodologies focus on the generation and variation of individual antibodies \cite{brownlee2011clever, tim2004ais, vikhar2016ea, kur2000imnet}. For instance, the Clonal Selection Algorithm (CLONALG) \cite{brownlee2007clonal, brownlee2011clever} involves repeated cloning and high-frequency mutation of the most promising antibodies to enhance solution quality. The Negative Selection Algorithm (NSA) \cite{forrest1994nsa, brownlee2011clever, ji2007nsa} emphasizes the identification and differentiation of normal and abnormal states to assess solution quality. Most of these AIS algorithms primarily concentrate on the changes within the antibodies themselves. 

Even though AIS algorithms possess many different varieties, most of the AIS algorithms are based on the core functionality shown in Algorithm \ref{alg_ais}. SAIS also grounded in these fundamental AIS principles. Within the main loop of the algorithm, AIS evaluates the affinity of the antibody population $P$ against each antigen $atg \in ATG$. Antibodies exhibiting the highest affinity undergo selection and mutation to generate new antibodies. This concept in AIS bears resemblance to the elite population strategy commonly employed in evolutionary algorithms \cite{brownlee2011clever, vikhar2016ea}. Ultimately, the algorithm yields a set of memory cells, representing an optimal or near-optimal solution to the problem. The 'memory' function of the AIS plays a crucial role similar to the human immune system, with each iteration enhancing the identification and improvement of antibody quality.

\begin{algorithm}
\caption{The Core Process of AIS Algorithms}
\label{alg_ais}
\begin{footnotesize}
\begin{algorithmic}[1] 
\State Initialise antibody population $P$
\State Define antigens $ATG$
\While{termination condition not met}
    \For{each antigen $atg$ in $ATG$}
        \State Evaluate affinity of $P$ to $atg$
        \State Select a subset of $P$ with highest affinity to $atg$
        \State Generate new antibodies by mutation
        \State Replace low affinity antibodies in $P$ with new antibodies
    \EndFor
    \State Update memory cells with high-affinity antibodies
\EndWhile
\State \textbf{return} memory cells
\end{algorithmic}
\end{footnotesize}
\end{algorithm}

An innovative approach that amalgamates AIS with the concept of symbiosis was first explored in \cite{hal2007symbais}, which led to the development of the Symbiotic Artificial Immune Systems with Partially Specified Antibodies (SymbAIS) algorithm, representing an advanced iteration of CLONALG \cite{brownlee2007clonal}. The core methodology of SymbAIS involves employing partially specified antibodies for the search process, progressively constructing categories of potential solutions, and iterative refining of these solutions until an optimal solution is identified. Comparative analyses reveal that while SymbAIS exhibits a longer duration in identifying the most suitable solution relative to the conventional clonal selection algorithm, it demonstrates superior efficacy in addressing problems characterized by a higher complexity of parameters. Notably, SymbAIS manifests its distinct advantages, particularly in scenarios where the CLONALG encounters limitations or fails to provide viable solutions.

Distinct from the SymbAIS algorithm, the Symbiotic Organisms Search (SOS) algorithm \cite{cheng2014sos} emerges as a unique meta-heuristic optimisation algorithm, diverging from the conventional AIS. SOS algorithm ingeniously leverages the three fundamental symbiotic relationships (mutualism, commensalism and parasitism) observed in the biological realm. The SOS algorithm creatively translates these symbiotic relationships into mathematical formulas (see Section \ref{subsect_mcp} for details) that encapsulate the intricate dynamics of symbiotic relationships between organisms. SOS algorithm is characterized by its minimal parameter requirements and rapid convergence in optimisation scenarios. Nonetheless, its efficacy can be influenced by the specific nature of the problem at hand and intrinsic attributes of the algorithm itself, such as susceptibility to local optima and the implications of population size. The empirical findings presented in \cite{cheng2014sos} indicate that the SOS algorithm demonstrates a broadly effective performance in processing benchmark functions characterized by multiple local optima. However, our investigations have identified a limitation of the SOS algorithm, particularly in scenarios involving large initial population sizes. In contrast, our SAIS algorithm demonstrates superior performance in effectively optimising a wide range of mathematical benchmarks under these conditions. This observation has led us to further explore the SAIS algorithm. In Section \ref{sect3}, we discuss the symbiotic relationships that are central to the SAIS algorithm and detail the core principles that guide its operation. Our approach addresses the question of why not simply reduce the size of the population by demonstrating that, in certain contexts, a larger population size can be leveraged for more effective optimisation of complex problems.





\section{Symbiotic Artificial Immune Systems} \label{sect3}

\subsection{Structure and Design of SAIS}
\begin{figure}[ht]
  \centering
  \includegraphics[width=0.7\linewidth]{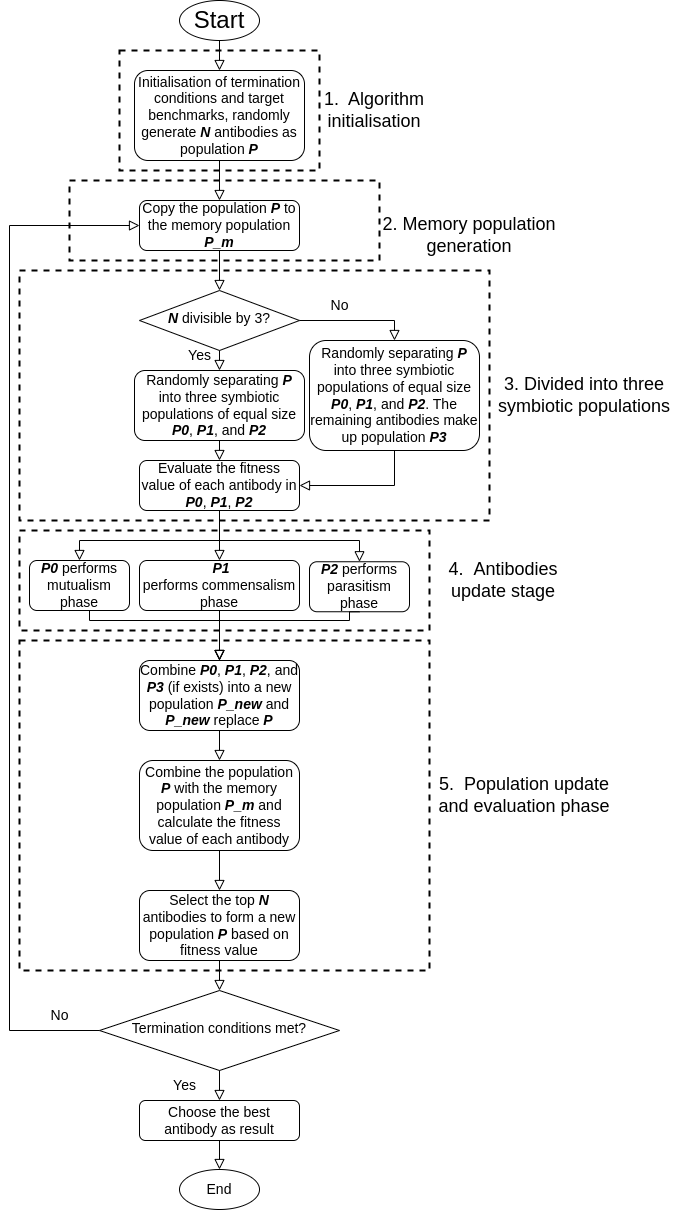}
  \caption{SAIS Algorithm Flow Chart}
  \Description{The main flow chart of SAIS Algorithm}
  \label{fig_flowchart}
\end{figure}

The flow chart of the Symbiotic Artificial Immune Systems algorithm is depicted in Figure \ref{fig_flowchart}. The portion enclosed within the dotted rectangle delineates the five pivotal steps that constitute the core of the SAIS algorithm. 

In the initial step of the SAIS, the termination conditions for the algorithm's iterations are established. This could involve setting an upper limit on the number of loops or determining whether the conditions for the target solution have been met. Additionally, during the initialization phase, SAIS randomly generates an initial population $P$ comprising $N$ antibodies, based on the known domain of the problem or function. 

In the second step, SAIS adopts the characteristics of AIS with memory cells. SAIS saves the state of the population before the algorithm is executed by copying/cloning the existing population $P$ to the new memory population $P_m$. 

As described by the first three dotted rectangles (three main step) at the end of the flow chart. SAIS has to not only retain the memory cells in AIS to judge the superior antibodies but also emulate the three symbiotic relationships in biology to promote antibody mutation. However, the SOS algorithm mentioned in the Section \ref{subsect_symbAIS_SOS} takes the same population step by step through the three different symbiotic relationships. SOS performs a mutualism phase followed by a commensalism phase and finally takes the population through a parasitism phase in the algorithmic cycle. This indicates that SOS applies all three symbiotic relationships of organisms to the same species. The fact is, SOS is not accurate in emulating the symbiotic relationships of organisms, as every organism in SOS will have all the three symbiotic relationships. In order to more accurately emulate symbiotic relationships in the biological area for better optimisation performance, SAIS will randomly divide the initial population into three sub-populations to perform mutualism, commensalism and parasitism (the realization of these three symbiotic relationships will be explained in detail in Section \ref{subsect_mcp}). If the size of this initial population is not divisible by three, the remaining individuals will remain in the initial population. SAIS partitions the population $P$ into three smaller sub-populations, facilitating concurrent symbiotic operations. These updated sub-populations are then merged and used to to replace the original population. The augmented population is subsequently integrated with the 'memory' population to produce a new population double the size of the original population. Thereafter, SAIS selects the top $N$ antibodies (ensuring that the population size is consistent in each iteration), constituting a new population via fitness evaluation. This procedure aids SAIS in preserving populations with lower fitness values for subsequent iterations.

In the field of AIS, the affinity value (fitness value) of an antibody is indicative of its binding efficacy to the antigen, serving as a measure of the antibody's quality \cite{tim2004ais}. Let $atb$ represent an antibody and $atg$ represent an antigen. The affinity (fitness) function $f$ of the antibody can be defined as follows:

\begin{equation} \label{eq_fi}
f(atb) = \text{function}(atb, atg)
\end{equation}
where the function quantifies the degree of binding between the antibody $ATB$ and the antigen $ATG$. In the experiments in the Section \ref{sect4}, our optimisation problems focused on locating the minimum value of the benchmark function. Consequently, the fitness value is expected to decrease progressively with each iteration until the specified condition is met.

\subsection{Mutualism, Commensalism and Parasitism} \label{subsect_mcp}
SAIS adopts the mathematical presentation of SOS \cite{cheng2014sos} for the process of Mutualism, Commensalism and Parasitism. However, SAIS is applied to the AIS algorithm in the form of parallel connection as shown in Figure \ref{fig_flowchart}.

In the mutualism phase, $atb_i$ is the i-th antibody in the mutualism population $P0$. Then another antibody $atb_j$ is randomly selected from the mutualism population $P0$ to interact with $atb_i$. Both antibodies are in a mutually beneficial relationship, with the goal of increasing the mutual survival advantage in the population. The update process can be represented as follows:
\begin{equation} \label{eq_mu}
\begin{aligned}
    atb_i' &= atb_i + \text{rand}(0, 1) \cdot (best\_atb - \mu \cdot b_f), \\
    atb_j' &= atb_j + \text{rand}(0, 1) \cdot (best\_atb - \mu \cdot b_f)
\end{aligned}
\end{equation}
where $atb_i'$ and $atb_j'$ are the updated antibodies, $best\_atb$ is the best antibody in the mutualism population, $\mu = \frac{atb_i + atb_j}{2}$ is the mutual factor, $b_f$ is a benefit factor randomly chosen from $\{1, 2\}$, and $\text{rand}(0, 1)$ generates a random number between 0 and 1.

In the commensalism phase, for each antibody $atb_i$ in the commensalism population $P1$, another antibody $atb_j$ is randomly selected from the same population. The update process can be represented as follows:
\begin{equation} \label{eq_co}
    atb_i' = atb_i + \text{rand}(-1, 1) \cdot (best\_atb - atb_j)
\end{equation}
where $atb_i'$ is the updated antibody, $best\_atb$ is the best antibody in the commensalism population, $atb_j$ is another randomly selected antibody, and $\text{rand}(-1, 1)$ generates a random number between -1 and 1. This phase aims to improve the fitness of each antibody $atb_i$ by leveraging the position of another antibody $atb_j$ in relation to the best-performing antibody in the population ($atb_j$ does not change in this phase).

In the parasitism phase, each antibody $atb_i$ in the parasitism population $P2$ undergoes a potential replacement. The process can be represented as follows:
\begin{equation} \label{eq_pa}
    atb_i' = \begin{cases} 
    new\_atb, & \text{if } f(new\_atb) \text{ better than } f(atb_i) \\
    atb_i, & \text{otherwise}
    \end{cases}
\end{equation}
where $atb_i'$ is the updated antibody, $new\_atb$ is a newly generated antibody, and $f$ is the fitness function. In this phase, a new antibody (parasite) replaces an existing one if it demonstrates superior fitness, where 'better' is defined according to the specific optimisation criteria of the problem (either maximization or minimisation).

\subsection{Pseudocode for SAIS}

\begin{algorithm}
\caption{SAIS Algorithm}
\label{alg_sais}
\begin{footnotesize}
\begin{algorithmic}[1]
\State Initialise population $P$ of $N$ antibodies
\State Determine globally optimal $global\_opt$
\While{termination condition not met}
    \State Copy $P$ to memory population $P_m$
    \State Divide $P$ into $P0$, $P1$, $P2$ populations
    
    \If{$len(P) \% 3 \neq 0$}
        \State Retain remainder in $P$
    \EndIf
    
    \State \textbf{Mutualism Phase:}
    \For{each antibodies $atb_i$ in $P0$}
        \State Select another antibody $atb_j$ randomly
        \State Perform mutualism equation \ref{eq_mu} on $atb_i$, $atb_j$
    \EndFor
    
    \State \textbf{Commensalism Phase:}
    \For{each antibody $atb_i$ in $P1$}
        \State Select another antibody $atb_j$ randomly
        \State Perform commensalism equation \ref{eq_co} on $atb_i$ using $atb_j$
    \EndFor
    
    \State \textbf{Parasitism Phase:}
    \For{each antibody $atb_i$ in $P2$}
        \State Generate a new antibody $new\_atb$
        \State Perform parasitism equation \ref{eq_pa} on $atb_i$ using $new\_atb$
    \EndFor
    
    \State Replace updated $P0$, $P1$, $P2$ into $P$
    \State $P \gets P + P_m$
    \State Sort $P$ by fitness and retain the best half
    \State $best\_fitness \gets$ best fitness value in $P$
    \If{$best\_fitness \text{ better than } global\_opt$}
        \State \textbf{break}
    \EndIf
\EndWhile
\State $best\_antibody \gets$ Antibody with best fitness in $P$
\State \Return $best\_antibody$, $best\_fitness$
\end{algorithmic}
\end{footnotesize}
\end{algorithm}

\begin{figure}[ht]
  \centering
  \includegraphics[width=0.7\linewidth]{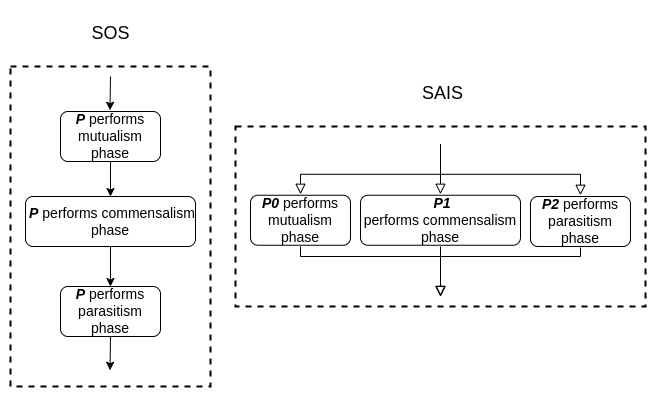}
  \caption{Comparison of Population Updating Methods in SAIS and SOS}
  \Description{Comparison between SAIS and SOS}
  \label{compare_sos_sais}
\end{figure}

The pseudocode of SAIS is shown in Algorithm \ref{alg_sais}. A population containing $N$ antibodies is first initialised. During the iterative process, the population will be randomly divided into three sub-populations of the same population size. These three sub-populations will undergo mutualism, commensalism and parasitism phases, respectively. If the population size is not divisible by three, the remaining antibodies are retained in the original population and no operation will be performed. In each phase, the antibodies are optimised by specific update rules. The algorithm continues to iterate until the termination condition is met, and finally returns the antibody with the best fitness. The innovative design of the SAIS enhances the AIS's capability to simulate symbiotic mechanisms found in biology with greater accuracy. As depicted in Figure \ref{compare_sos_sais}, SAIS concurrently processes the three symbiotic relationships (mutualism, commensalism, and parasitism) corresponding to population update methods within biology. This parallel processing approach allows each subpopulation to use a different update method at the same time, which is different from the sequential step-by-step execution of population $P$ in SOS. Consequently, SAIS's concurrent execution, demonstrated by the simultaneous mutualism, commensalism, and parasitism phases in subpopulations $P0$, $P1$, and $P2$, not only significantly enhances population diversity but also reduces the time to completion. Such a parallel structure is particularly beneficial when handling large-scale and complex tasks, as the capacity to perform multiple operations concurrently is crucial for optimising computational resource utilisation.

\section{Experiments} \label{sect4}

\subsection{The Analysis of Max Iteration and Population Size} \label{subsect_analysis_max_size}
In Section \ref{sect3}, we have presented SAIS in detail, with a specific emphasis on population size and iteration number as critical parameters. Our preliminary experiments revealed performance variations in SAIS under different parameter settings. To explore the selection of suitable values for these parameters, we initially chose to conduct an analysis based on 26 benchmark functions. The 26 benchmark functions initially employed in \cite{dervis2009abc} to evaluate the efficacy of algorithms like GA \cite{brownlee2011clever}, DE \cite{dervis2009abc} and PSO \cite{ken1995pso} have been widely recognized for their comprehensiveness. Further scrutiny by \cite{cheng2012pba, cheng2014sos} involved testing SOS and Particle Bee Algorithm (PBA) \cite{cheng2012pba} against these benchmark functions, yielding positive outcomes. Encompassing a diverse range of problem types including uni-modal, multi-modal, and multidimensional features, these benchmark functions are instrumental in delineating the scope of problems an algorithm can adeptly address, thereby providing robust experiments for SAIS. Table \ref{tab:pop50} contains the details of these 26 benchmark functions, including their names and optimal solutions. Our primary objective was to assess the performance of SAIS and determine the required number of iterations and population size for achieving optimal solutions.

The benchmark study conducted in \cite{cheng2014sos} fixed the population size ($p$) at $50$ and the maximum iteration number ($i$) at $500,000$. To ensure a fair comparison among different combinations of $p$ and $i$, we maintained a constant maximum computational cost ($B$) for each combination, which was set to $B = p \cdot i = 2.5 \times 10^7$. Consequently, we established specific experimental settings, as outlined in Table \ref{tab:expdesign}. We chose to increase the population size ($p$) from $50$ to $50,000$ because we observed that SAIS tends to perform better with larger population sizes.

Table \ref{tab:iteration_and_sr} presents the results for four different combinations of $p$ and $i$. It is worth noting that, for each parameter combination, we conducted $30$ trials of SAIS on a single problem to ensure the robustness and statistical significance of the results. We calculated the success rate, which serves as an evaluation metric for SAIS performance. The success rate is defined as the number of times, out of $30$ trials, that SAIS successfully found the optimal solutions.

In addition, we employed the iteration mean to measure the efficiency of SAIS configured with different parameters, but only when SAIS successfully found the optimal solutions. To clarify, those trials in which SAIS did not find the optimal solutions after reaching the preset iteration limit were excluded from this calculation. Over the process of $30$ trials for a single benchmark problem, SAIS might attain the optimal solution using varying numbers of iterations. While the four settings inherently correspond to four different values of $i$, our goal was to examine whether the required number of iterations decreased as the population size increased. The iteration STD represents the standard deviation of the required iterations to find the optimal solutions.

In Table \ref{tab:iteration_and_sr}, we observed that SAIS performed best with $p=50,000$ and $i=500$, achieving optimal solutions for $22$ out of $26$ tasks with a $100\%$ success rate. In comparison, SAIS with $p=5,000$ outperformed the other two configurations, while SAIS with $p=50$ demonstrated the worst performance. Moreover, as we increased the value of $p$, we noted a corresponding improvement in the success rate. For example, on benchmark problems $11$ and $13$, SAIS achieved a $100\%$ success rate only with $p=50,000$, demonstrating a clear monotonic increase in success rate from $p=50$ to $p=50,000$. Therefore, based on these $26$ benchmark problems, it can be concluded that, overall, SAIS tends to perform better when configured with a larger population size to achieve promising results. However, there is an exception in the case of benchmark problem 16, where the success rate increased from $46.67\%$ to $100\%$ as $p$ increased from $500$ to $5,000$, while SAIS with the other two settings obtained 0\%. This underscores the importance of carefully considering the choice of $p$ and $i$ as these parameters play a crucial role in SAIS performance and may vary depending on the specific problem.

In terms of the iteration mean, the required number of iterations to reach convergence decreases overall as the population size increases. However, it is worth noting that the actual computational cost increases, even when we assign an equal value of $B$ to all four situations. For example, in benchmark problem 1, SAIS with $p=50,000$ takes an average of 23. 23 iterations less than the other three situations to find optimal solutions, but the actual computational cost is higher compared to SAIS with $p=500$ and $p=5,000$. It is advisable to use a large population size when dealing with specific and unfamiliar problems, as it ensures better performance at the expense of computational cost. Furthermore, in Table \ref{tab:iteration_and_sr}, for benchmark problems 20, 24, 25, and 26, SAIS with larger population sizes inversely requires more iterations to find optimal solutions, possibly due to the characteristics of these benchmark problems. Therefore, SAIS may exhibit lower efficiency when configured with a larger population size, but it consistently achieves a higher success rate and maintains stable performance.

\begin{table}
\centering
\begin{tabular}{c|cc}
  & $p$     & $i$      \\ 
\hline
1 & $50$    & $500,000$  \\
2 & $500$   & $50,000$   \\
3 & $5,000$  & $5,000$    \\
4 & $50,000$ & $500$     \\
\hline
\end{tabular}
\caption{The summary of the experiment design with an aim to investigate the performance of SAIS in relation to various selections of population size and iteration.}
\label{tab:expdesign}
\end{table}

\begin{table*}[htb]
\centering
\resizebox{12cm}{!}{
\begin{tabular}{c|cccc|cccc|cccc} 
\hline
                  & \multicolumn{4}{c|}{Success rate (\%)} & \multicolumn{4}{c|}{Iteration Mean}  & \multicolumn{4}{c}{Iteration STD}  \\ 
\hline
Benchmark Problem & 50     & 500    & 5000   & 50000       & 50       & 500     & 5000   & 50000  & 50       & 500    & 5000  & 50000  \\ 
\hline
1                 & 0.00   & 96.67  & 100.00 & 100.00      & n/a      & 33.00   & 27.70  & 23.23  & n/a      & 4.38   & 3.05  & 2.21   \\
2                 & 13.33  & 100.00 & 100.00 & 100.00      & 10966.50 & 35.77   & 32.13  & 27.50  & 13995.32 & 4.35   & 2.92  & 1.63   \\
3                 & 100.00 & 100.00 & 100.00 & 100.00      & 25.27    & 19.90   & 17.73  & 15.43  & 3.34     & 1.67   & 1.05  & 1.28   \\
4                 & 100.00 & 100.00 & 100.00 & 100.00      & 28.07    & 24.90   & 23.30  & 21.47  & 3.10     & 1.35   & 1.47  & 1.04   \\
5                 & 0.00   & 83.33  & 100.00 & 100.00      & n/a      & 37.84   & 29.63  & 27.03  & n/a      & 5.82   & 2.89  & 2.28   \\
6                 & 30.00  & 100.00 & 100.00 & 100.00      & 132.44   & 11.87   & 8.93   & 6.40   & 200.69   & 2.50   & 2.27  & 1.65   \\
7                 & 100.00 & 100.00 & 100.00 & 100.00      & 41.10    & 26.83   & 24.13  & 22.60  & 19.99    & 2.56   & 0.86  & 1.28   \\
8                 & 43.33  & 96.67  & 100.00 & 100.00      & 22667.08 & 31.10   & 26.37  & 22.63  & 41911.18 & 4.14   & 2.89  & 2.79   \\
9                 & 100.00 & 100.00 & 100.00 & 100.00      & 24.80    & 23.20   & 20.97  & 19.53  & 2.14     & 1.52   & 1.27  & 0.82   \\
10                & 100.00 & 100.00 & 100.00 & 100.00      & 31.60    & 26.37   & 24.43  & 22.63  & 4.97     & 2.11   & 1.45  & 1.22   \\
11                & 36.67  & 80.00  & 93.33  & 100.00      & 136.91   & 27.92   & 21.75  & 12.33  & 179.66   & 22.62  & 9.20  & 4.38   \\
12                & 0.00   & 0.00   & 26.67  & 90.00       & n/a      & n/a     & 117.25 & 100.41 & n/a      &        & 18.41 & 12.93  \\
13                & 0.00   & 16.67  & 76.67  & 100.00      & n/a      & 91.00   & 71.52  & 59.70  & n/a      & 11.77  & 5.78  & 4.63   \\
14                & 100.00 & 100.00 & 100.00 & 100.00      & 77.40    & 80.40   & 79.93  & 76.77  & 5.55     & 2.14   & 1.11  & 0.68   \\
15                & 0.00   & 0.00   & 3.33   & 6.67        & n/a      & n/a     & 305.00 & 256.00 & n/a      & n/a    & n/a   & 1.41   \\
16                & 0.00   & 46.67  & 100.00 & 0.00        & n/a      & 1098.71 & 668.07 & n/a    & n/a      & 132.97 & 19.82 & n/a    \\
17                & 100.00 & 100.00 & 100.00 & 100.00      & 89.13    & 105.90  & 105.23 & 100.33 & 3.17     & 1.42   & 0.86  & 0.76   \\
18                & 100.00 & 100.00 & 100.00 & 100.00      & 84.20    & 99.40   & 98.40  & 93.93  & 2.81     & 1.35   & 0.89  & 0.52   \\
19                & 0.00   & 0.00   & 0.00   & 0.00        & n/a      & n/a     & n/a    & n/a    & n/a      & n/a    & n/a   & n/a    \\
20                & 100.00 & 100.00 & 100.00 & 100.00      & 140.50   & 177.17  & 180.03 & 172.63 & 3.21     & 1.78   & 1.27  & 0.81   \\
21                & 100.00 & 100.00 & 100.00 & 100.00      & 94.80    & 111.87  & 110.97 & 105.90 & 2.46     & 1.25   & 0.72  & 0.55   \\
22                & 0.00   & 0.00   & 0.00   & 0.00        & n/a      & n/a     & n/a    & n/a    & n/a      & n/a    & n/a   & n/a    \\
23                & 0.00   & 0.00   & 0.00   & 0.00        & n/a      & n/a     & n/a    & n/a    & n/a      & n/a    & n/a   & n/a    \\
24                & 100.00 & 100.00 & 100.00 & 100.00      & 92.50    & 117.10  & 128.33 & 136.87 & 4.22     & 6.70   & 5.66  & 9.10   \\
25                & 100.00 & 100.00 & 100.00 & 100.00      & 92.47    & 108.47  & 107.90 & 103.73 & 2.70     & 1.61   & 1.65  & 1.96   \\
26                & 100.00 & 100.00 & 100.00 & 100.00      & 145.63   & 174.90  & 174.20 & 166.50 & 3.47     & 1.99   & 0.85  & 0.82   \\
\hline
\end{tabular}}
\caption{The experimental results of investigating the correlation between the performance of SAIS and parameter selection.}
\label{tab:iteration_and_sr}
\end{table*}

\subsection{SAIS on 26 Benchmark Unconstrained Mathematical Problems}
Based on the analysis results obtained from Section \ref{subsect_analysis_max_size}, we have opted to use $p=50,000$ and $i=500$ for SAIS for the purpose of comparison with other approaches discussed in \cite{cheng2014sos}. All the selected approaches in \cite{cheng2014sos} are listed in Table \ref{tab:pop50} and are configured with $p=50$ and $i=500,000$. Although the settings differ, we have ensured that the value of $B$ remains consistent to maintain experimental fairness. It's important to note that our proposed SAIS typically employs a larger value of $p$, setting it apart from the other approaches. In our experiments, as shown in Table \ref{tab:pop50}, each method has been allocated the same maximum computational cost and run 30 times for each problem.

In Table \ref{tab:pop50}, the ``Min" column contains 26 optimal solutions for the corresponding problems. The ``Mean" and "StdDev" columns respectively represent the average values that each approach obtained in 30 runs and the standard deviation for each problem. Overall, SAIS achieved results comparable to SOS, with both approaches successfully finding the global minimum in 20 and 21 out of 26 tasks, respectively. It is worth noting that SOS was initially reported to have solved 22 problems, but upon reproducing the experiments, we discovered that SOS could only solve 21 problems. In Table \ref{tab:pop50}, we have highlighted in red the discrepancies observed in our replication experiments. Specifically, for the SOS applied to benchmark functions 22 and 23, the results did not align with those reported in \cite{cheng2014sos}. In terms of computational cost, in theory, SAIS has an advantage over SOS. In SOS, the population goes through three phases to update solutions, and in one iteration, each phase requires a round of evaluation on the whole population. Meanwhile, due to the design of SAIS, the population has been randomly divided into three groups for evolving the promising solutions in a single iteration, with each group undergoing different evolutionary operators (Mutualism, Commensalism and Parasitism). Consequently, the computational cost of SOS is three times higher than that of SAIS. We believe this difference makes SAIS stand out, especially when applied to solving deep learning-related problems (e.g., hyperparameter optimisation), as the evaluation of a deep learning model is a very expensive task \cite{yuan2021systematic} \cite{yuan2021genetic} \cite{yuan2021novel}.

Additionally, in the case of Benchmark 23, we observed that SAIS outperformed other approaches, while most of them remained trapped at the local optimal point of $0.66667$. Although SAIS did not reach the global optimum, it demonstrated superior performance compared to the other methods. Furthermore, regarding Benchmark 12, even though SAIS did not reach the optimum, it came very close. In our experiments, we set the convergence criterion to require results to match to at least 12 decimal places. In contrast, for Benchmark 5, SOS is further from the global minimum.

\begin{table*}[htb]
\centering
\resizebox{10cm}{!}{
\begin{tabular}{|l|l|r|r|r|r|r|r|r|r|}
\hline
\textbf{Functions} & \textbf{Metric} & \textbf{Min} & \textbf{GA} & \textbf{DE} & \textbf{PSO} & \textbf{BA} & \textbf{PBA} & \textbf{SOS} & \textbf{SAIS} \\
\hline
1. Beale & Mean & 0 & 0 & 0 & 0 & 1.88E-05 & 0 & 0 & 0 \\
 & StdDev &  & 0 & 0 & 0 & 1.94E-05 & 0 & 0 & 0 \\
2. Easom & Mean & -1 & -1 & -1 & -1 & -0.99994 & -1 & -1 & -1 \\
 & StdDev &  & 0 & 0 & 0 & 4.50E-05 & 0 & 0 & 0 \\
3. Matyas & Mean & 0 & 0 & 0 & 0 & 0 & 0 & 0 & 0 \\
 & StdDev &  & 0 & 0 & 0 & 0 & 0 & 0 & 0 \\
4. Bohachevsky1 & Mean & 0 & 0 & 0 & 0 & 0 & 0 & 0 & 0 \\
 & StdDev &  & 0 & 0 & 0 & 0 & 0 & 0 & 0 \\
5. Booth & Mean & 0 & 0 & 0 & 0 & 0.00053 & 0 & 0.03382 & 0 \\
 & StdDev &  & 0 & 0 & 0 & 0.00074 & 0 & 0.1287 & 0 \\
6. Michalewicz2 & Mean & -1.8013 & -1.8013 & -1.8013 & -1.57287 & -1.8013 & -1.8013 & -1.8013 & -1.8013 \\
 & StdDev &  & 0 & 0 & 0.11986 & 0 & 0 & 0 & 0 \\
7. Schaffer & Mean & 0 & 0.00424 & 0 & 0 & 0 & 0 & 0 & 0 \\
 & StdDev &  & 0.00476 & 0 & 0 & 0 & 0 & 0 & 0 \\
8. Six Hump Camel Back & Mean & -1.03163 & -1.03163 & -1.03163 & -1.03163 & -1.03163 & -1.03163 & -1.03163 & -1.03163 \\
 & StdDev &  & 0 & 0 & 0 & 0 & 0 & 0 & 0 \\
9. Boachevsky2 & Mean & 0 & 0.06829 & 0 & 0 & 0 & 0 & 0 & 0 \\
 & StdDev &  & 0.07822 & 0 & 0 & 0 & 0 & 0 & 0 \\
10. Boachevsky3 & Mean & 0 & 0 & 0 & 0 & 0 & 0 & 0 & 0 \\
 & StdDev &  & 0 & 0 & 0 & 0 & 0 & 0 & 0 \\
11. Shubert & Mean & -186.73 & -186.73 & -186.73 & -186.73 & -186.73 & -186.73 & -186.73 & -186.73 \\
 & StdDev &  & 0 & 0 & 0 & 0 & 0 & 0 & 0 \\
12. Colville & Mean & 0 & 0.01494 & 0.04091 & 0 & 1.1176 & 0 & 0 & 2.79E-08 \\
 & StdDev &  & 0.00736 & 0.08198 & 0 & 0.46623 & 0 & 0 & 1.50E-07 \\
13. Michalewicz5 & Mean & -4.6877 & -4.64483 & -4.68348 & -2.49087 & -4.6877 & -4.6877 & -4.6877 & -4.6877 \\
 & StdDev &  & 0.09785 & 0.01253 & 0.25695 & 0 & 0 & 0 & 0 \\
14. Zakharov & Mean & 0 & 0.01336 & 0 & 0 & 0 & 0 & 0 & 0 \\
 & StdDev &  & 0.00453 & 0 & 0 & 0 & 0 & 0 & 0 \\
15. Michalewicz10 & Mean & -9.6602 & -9.49683 & -9.59115 & -4.00718 & -9.6602 & -9.6602 & -9.65982 & -9.42346 \\
 & StdDev &  & 0.14112 & 0.06421 & 0.50263 & 0 & 0 & 0.00125 & 0.25186 \\
16. Step & Mean & 0 & 1.17E+03 & 0 & 0 & 5.1237 & 0 & 0 & 1.19E-10 \\
 & StdDev &  & 76.56145 & 0 & 0 & 0.39209 & 0 & 0 & 5.78E-11 \\
17. Sphere & Mean & 0 & 1.11E+03 & 0 & 0 & 0 & 0 & 0 & 0 \\
 & StdDev &  & 74.21447 & 0 & 0 & 0 & 0 & 0 & 0 \\
18. SumSquares & Mean & 0 & 1.48E+02 & 0 & 0 & 0 & 0 & 0 & 0 \\
 & StdDev &  & 12.40929 & 0 & 0 & 0 & 0 & 0 & 0 \\
19. Quartic & Mean & 0 & 0.1807 & 0.00136 & 0.00116 & 1.72E-06 & 0.00678 & 9.13E-05 & 8.86E-05\\
 & StdDev &  & 0.02712 & 0.00042 & 0.00028 & 1.85E-06 & 0.00133 & 3.71E-05 & 1.29E-05\\
20. Schwefel 2.22 & Mean & 0 & 11.0214 & 0 & 0 & 0 & 7.59E-10 & 0 & 0 \\
 & StdDev &  & 1.38686 & 0 & 0 & 0 & 7.10E-10 & 0 & 0 \\
21. Schwefel 1.2 & Mean & 0 & 7.40E+03 & 0 & 0 & 0 & 0 & 0 & 0 \\
 & StdDev &  & 1.14E+03 & 0 & 0 & 0 & 0 & 0 & 0 \\
22. Rosenbrock & Mean & 0 & 1.96E+05 & 18.20394 & 15.088617 & 28.834 & 4.2831 & 1.04E-07 \color{red}{(0.9335)} & 26.40837\\
 & StdDev &  & 3.85E+04 & 5.03619 & 24.170196 & 0.10597 & 5.7877 & 2.95E-07 \color{red}{(1.2681)} & 0.16131 \\
23. Dixon Price & Mean & 0 & 1.22E+03 & 0.66667 & 0.66667 & 0.66667 & 0.66667 & 0  \color{red}{(0.66667)} & 0.5 \\
 & StdDev &  & 2.66E+02 & E-9 & E-8 & 1.16E-09 & 5.65E-10 & 0 & 0 \\
24. Rastrigin & Mean & 0 & 52.92259 & 11.71673 & 43.97714 & 0 & 0 & 0 & 0 \\
 & StdDev &  & 4.56486 & 2.53817 & 11.72868 & 0 & 0 & 0 & 0 \\
25. Griewank & Mean & 0 & 10.63346 & 0.00148 & 0.01739 & 0 & 0.00468 & 0 & 0 \\
 & StdDev &  & 1.16146 & 0.00296 & 0.02081 & 0 & 0.00672 & 0 & 0 \\
26. Ackley & Mean & 0 & 14.67178 & 0 & 0.16462 & 0 & 3.12E-08 & 0 & 0 \\
 & StdDev &  & 0.17814 & 0 & 0.49387 & 0 & 3.98E-08 & 0 & 0 \\
\hline
\multicolumn{2}{|l|}{Count of algorithm found global minimum} &  & 9 & 18 & 17 & 18 & 20 & 22 \color{red}{(21)} & 20\\
\hline
\end{tabular}}
\caption{Algorithm experiment results, the definitions of 26 benchmark functions can be found in \cite{cheng2014sos}. It should be noted that we have replicated all SOS experiments and have identified differences between the results reported in the original paper and our experimental findings, which are highlighted.}
\label{tab:pop50}
\end{table*}

\subsection{Comparison of SAIS with other AIS}
Building upon the innovative concept of our proposed SAIS inspired by the immunology system, we sought to benchmark its efficacy against other AIS algorithms: CLONALG \cite{brownlee2007clonal} and NSA \cite{forrest1994nsa, ji2007nsa}. We have opted to use $p=50,000$ and $i=500$ for SAIS for the purpose of comparison with other AIS algorithms. To ensure a fair and rigorous comparison, we adhered to parameter configurations for competing algorithms as delineated in \cite{brownlee2011clever, brownlee2007clonal, ji2007nsa}. The initial population size of both GA and NSA in this experiment is $50$. GA and NSA allocated the same maximum number of iterations $50,000$ and ran $30$ times for each benchmark. This setup allows us to ensure that the value of $B$ is consistent in each different experiment to maintain the fairness of the experiments.

Table \ref{tab:comparisonAIS} presents the outcome of our experiments. A cursory glance reveals the distinct advantage SAIS holds across multiple benchmarks. Notably, in benchmark functions where traditional AIS algorithms such as CLONALG and NAS exhibit variability in performance, SAIS consistently achieves lower mean values, signifying superior optimisation capabilities. Furthermore, the standard deviation (StdDev) associated with SAIS results is significantly lower, indicating a higher degree of stability in finding optimal solutions.

In Benchmarks 1, 2, 3 and 4, the mean fitness values of SAIS achieved were in the order of $10^{-13}$. For instance, in Benchmark 1, SAIS achieved a mean fitness of $4.07E{-13}$ with a standard deviation of $3.25E{-13}$, and in Benchmark 2, SAIS achieved a perfect mean fitness of $-1$ with a nearly zero standard deviation of $2.67E{-13}$.

In Benchmark 22 (Rosenbrock), SAIS significantly outperformed both CLONALG and NAS, achieving a mean fitness value of $26.4083792$ with a standard deviation of $0.16131842$. Although the mean fitness indicates a departure from the near-zero values observed in other benchmarks, the relative comparison with CLONALG and NAS demonstrates SAIS's superior handling of Benchmark 22's notorious complexity and its propensity for local optima. Benchmark 23 (Dixon-Price) presented an intriguing scenario where SAIS attained a mean fitness of $0.5$ with a standard deviation of $0$, indicating perfect convergence to a solution. Considering Benchmark 23 as a thirty-dimensional benchmark, the performance of SAIS contrasts with that of the other two AIS algorithms. And we can find that SAIS presents an absolute advantage in handling all these thirty-dimensional benchmarks from 16 to 26. This highlights the adaptability and efficiency of SAIS in optimising high-dimensional problems. Through the benchmark functions with uni-modal characteristics such as Benchmarks 1, 2, 3, 14, 20, 21, etc., SAIS presents the best adaptation and the most stable standard deviation among the three algorithms, which illustrates the characteristics of SAIS itself in terms of its high robustness.

These observations highlight several key advantages of SAIS: the high accuracy achieved in various benchmarks, the robustness in complex optimisation environments. In previous comparisons with other non-AIS algorithms and with other AIS algorithms, we can identify the precision of SAIS in solving complex optimisation problems. In Section \ref{sect5} we will delve into the role of different symbiotic relationships in SAIS and compare it with ablation experiments.

\begin{table}[htb]
\centering
\resizebox{6cm}{!}{
\begin{tabular}{c|cc|cc|cc} 
\hline
          & \multicolumn{2}{c|}{CLONALG} & \multicolumn{2}{c|}{NAS} & \multicolumn{2}{c}{SAIS}  \\ 
\hline
Benchmark & Mean       & StdDev         & Mean       & StdDev      & Mean       & StdDev      \\ 
\hline
1         & 0.17781805 & 0.32231834     & 1.69E-06   & 1.61E-06    & 4.07E-13   & 3.25E-13    \\
2         & -0.9999981 & 1.45E-06       & -0.9991504 & 0.00080705  & -1         & 2.67E-13    \\
3         & 9.95E-08   & 8.71E-08       & 5.21E-07   & 5.60E-07    & 5.34E-13   & 3.14E-13    \\
4         & 2.53E-05   & 3.02E-05       & 0.01145236 & 0.00782652  & 4.51E-13   & 2.70E-13    \\
5         & 0.33816697 & 0.23910833     & 6.41E-05   & 6.45E-05    & 3.42E-13   & 2.55E-13    \\
6         & -1.8012783 & 2.74E-05       & -1.8012998 & 3.67E-06    & -1.8013023 & 9.24E-07    \\
7         & 7.71E-07   & 6.21E-07       & 0.00056121 & 0.00051364  & 4.34E-13   & 2.90E-13    \\
8         & -1.031624  & 4.57E-06       & -1.0316254 & 2.35E-06    & -1.0316285 & 3.19E-13    \\
9         & 1.23E-06   & 1.23E-06       & 0.00108329 & 0.00105122  & 3.93E-13   & 2.90E-13    \\
10        & 6.24E-06   & 5.93E-06       & 0.00368062 & 0.00348445  & 4.01E-13   & 3.02E-13    \\
11        & -186.72915 & 0.00172978     & -186.73007 & 0.00064008  & -186.73061 & 0.00025162  \\
12        & 0.06594855 & 0.03271116     & 0.98124422 & 0.53639183  & 2.80E-08   & 1.51E-07    \\
13        & -4.2838533 & 0.13369352     & -4.4518724 & 0.08294569  & -4.6876581 & 4.40E-08    \\
14        & 0.45738845 & 0.09352363     & 9.62588572 & 2.02332255  & 7.29E-13   & 1.47E-13    \\
15        & -5.9811174 & 0.26750757     & -6.6448745 & 0.28594838  & -9.4234669 & 0.25186417  \\
16        & 7.66642108 & 0.76361459     & 67.8664793 & 5.45301191  & 1.20E-10   & 5.79E-11    \\
17        & 7.50320437 & 0.62416044     & 25566.1748 & 2195.2511   & 8.32E-13   & 1.07E-13    \\
18        & 95.672352  & 14.3965129     & 2898.3747  & 264.168392  & 7.80E-13   & 1.34E-13    \\
19        & 45.0154933 & 8.05428601     & 15.212667  & 2.25707204  & 8.86E-05   & 1.30E-05    \\
20        & 15.1282702 & 17.4169497     & 81.3986359 & 9.52909104  & 8.99E-13   & 6.02E-14    \\
21        & 101.165018 & 9.19969676     & 295078.108 & 24316.8356  & 8.39E-13   & 1.00E-13    \\
22        & 1382.9522  & 689.132506     & 39026615.6 & 3193770.32  & 26.4083792 & 0.16131842  \\
23        & 340.203437 & 63.1686022     & 217674.762 & 44274.2452  & 0.5        & 0           \\
24        & 203.690271 & 13.5334897     & 268.288934 & 7.90957871  & 7.75E-13   & 1.18E-13    \\
25        & 0.31171311 & 0.02257614     & 229.772105 & 20.2426248  & 7.99E-13   & 1.08E-13    \\
26        & 13.8488845 & 7.78037007     & 18.5381361 & 0.28837621  & 8.92E-13   & 6.17E-14    \\
\hline
\end{tabular}}
\caption{The experimental results of comparing SAIS with CLONALG and NAS.}
\label{tab:comparisonAIS}
\end{table}

\section{Ablation Study} \label{sect5}
Ablation study is important for understanding the inner workings of SAIS and for efficiency studies of the algorithm. We elucidate their role by systematically removing three different symbiotic operators (Mutualism, Commensalism and Parasitism) of SAIS and by running $30$ experiments on the feature benchmark. 

\begin{figure}[ht]
    \centering
    \includegraphics[width=0.75\linewidth]{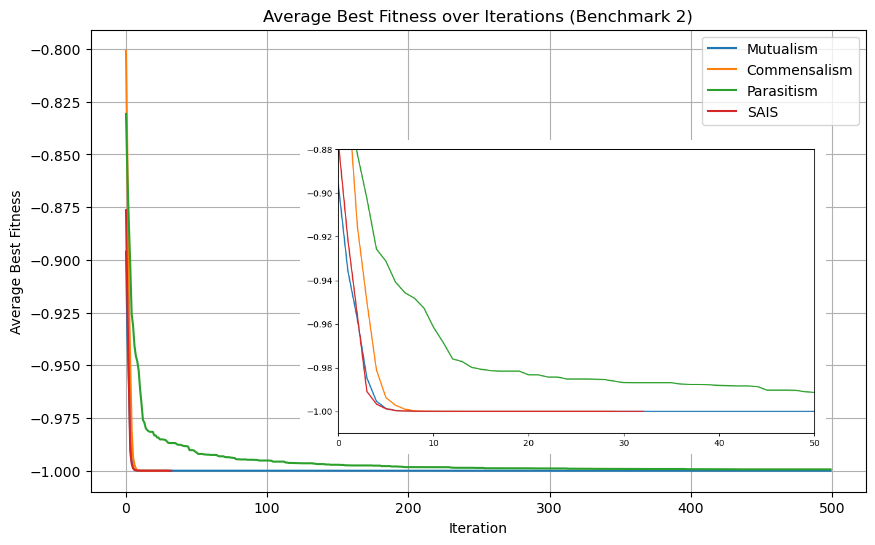}
    \caption{The change in the average fitness values of four models over 30 experimental runs on Benchmark Easom}
    \label{fig:bm2}
\end{figure}

\begin{figure}[ht]
    \centering
    \includegraphics[width=0.75\linewidth]{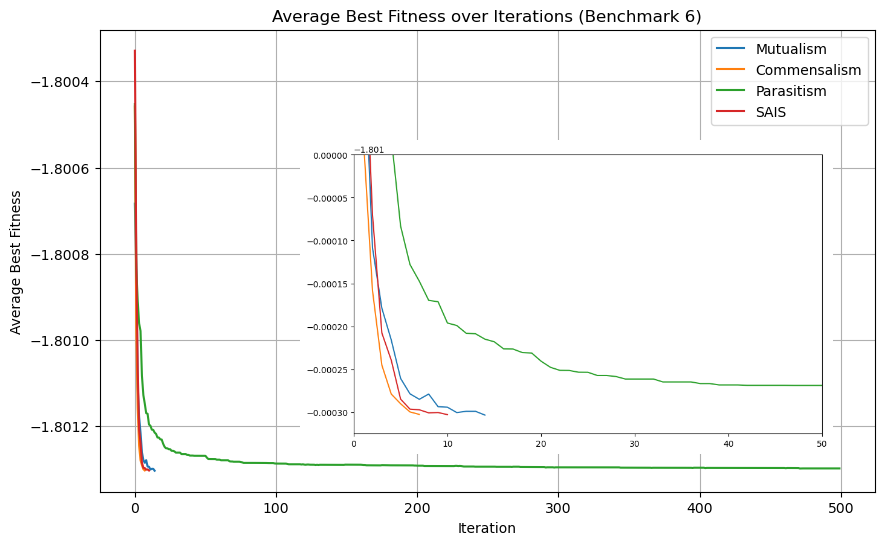}
    \caption{The change in the average fitness values of four models over 30 experimental runs on Benchmark Michalewicz2}
    \label{fig:bm6}
\end{figure}

\begin{figure}[ht]
    \centering
    \includegraphics[width=0.75\linewidth]{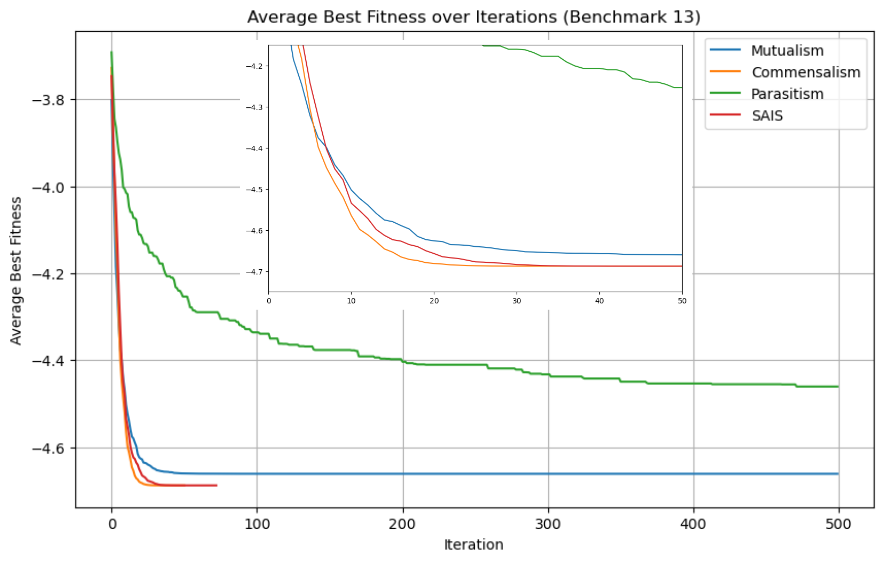}
    \caption{The change in the average fitness values of four models over 30 experimental runs on Benchmark Michalewicz5}
    \label{fig:bm13}
\end{figure}

\begin{figure}[ht]
    \centering
    \includegraphics[width=0.75\linewidth]{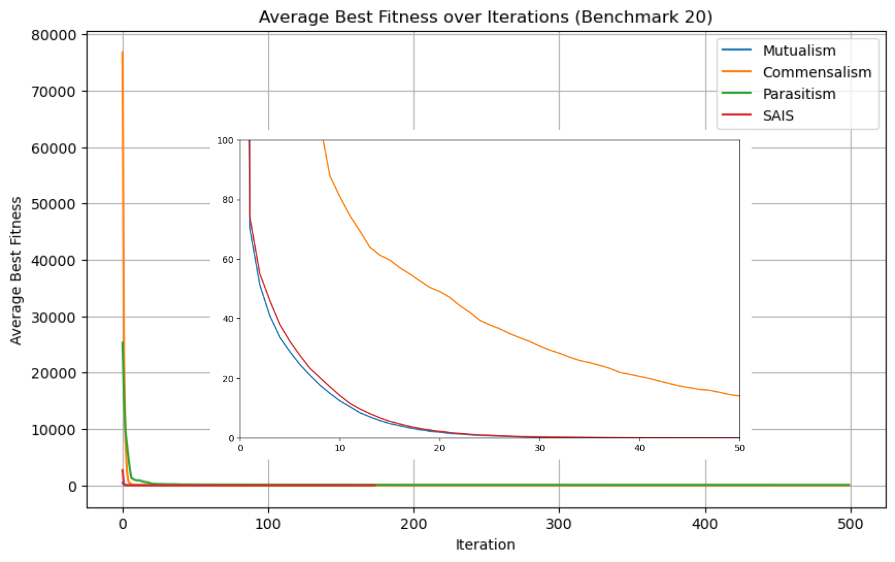}
    \caption{The change in the average fitness values of four models over 30 experimental runs on Benchmark Schwefel 2.22}
    \label{fig:bm20}
\end{figure}

We have selected four benchmark functions for our ablation study by analysing Table \ref{tab:iteration_and_sr} respectively. Among them, Benchmark 6 and 20 represent the performance of SAIS on the basis of a population of $50000$ for the least number of iterations and the most number of iterations, respectively. Benchmark $2$ is selected empirically, as we observed some interesting phenomena of SAIS based on this benchmark question. For a more comprehensive analysis, we searched for Benchmark 13 as part of our ablation experiments using a calculation of the median of the average number of iterations for a population of $50,000$ over the 26 benchmark functions. Figures \ref{fig:bm2}, \ref{fig:bm6}, \ref{fig:bm13}, \ref{fig:bm20} show the change in average fitness over $500$ iterations for each of our four models with different symbiotic operators.

The horizontal coordinates of all four graphs indicate the average number of iterations for each algorithm. The vertical coordinate represents the average of the best fitness of the iterative process for each algorithm over $30$ experiments. And a zoomed-in line graph at $50$ iterations is placed in each image for ease of observation. The four different coloured lines represent the four different algorithms. SAIS denotes the algorithm with all symbiotic operators. The remaining three represent algorithms that contain only one symbiotic operator. It can be seen from these figures that the algorithms that reach $500$ iterations are those that do not find the global optimal solution and exceed our preset number of iterations. It is worth noting that each line represents the average value of the fitness and the point of convergence is different on each fold. For example, some algorithms that require only $15$ iterations to find the optimal antibody in the first experiment may require $17$ iterations to find the optimal antibody in the second experiment. Where the fitness of the first 15 iterations will be the average of the results of all runs up to 15 iterations included. The next two iterations will be the average of all runs up to 16 and 17 iterations. The algorithm stops as soon as it finds the optimal solution for the benchmark function, so except for the algorithm containing only $Parasitism$, which did not succeed in finding a solution for any of the four problems, the other three algorithms each demonstrated strengths for a different benchmark function.

A comparison of these experiments shows that SAIS converges to the optimal solution of the function on all four benchmark functions. Algorithms containing $Commensalism$ only were unable to find the optimal antibody on Benchmark 20. The algorithm containing only $Mutualism$ symbiotic operators is unable to find the optimal solution on Benchmark 13 and the final average fitness function only converges to $4.6609$, which clearly shows that the speed and the number of iterations for finding the optimal antibody in Benchmark 2, 6, and 13 are not as good as that of SAIS.

\section{Conclusions} \label{sect6}
In this paper, we propose a new AIS algorithm inspired by symbiotic relationships in biology. Our experimental results show that SAIS can achieve excellent results in optimization problems. We experimentally investigated and found that SAIS tends to perform better with a larger population size. Additionally, in an ablation study, we discovered that SAIS with three symbiotic operators simultaneously outperforms the algorithm with only one of them in handling the benchmark function. In our ablation experiments, SAIS is the only algorithm that successfully finds the optimal solution for the selected benchmark function. We hope that SAIS can pave the way for new developments in bio-inspired and immune-inspired computing.


\bibliographystyle{ACM-Reference-Format}
\bibliography{sample-base}

\appendix

\end{document}